\documentclass{article}
\usepackage{ijcai24}

\usepackage{times}
\usepackage{soul}
\usepackage{url}
\usepackage[hidelinks]{hyperref}
\usepackage[utf8]{inputenc}
\usepackage[small]{caption}
\usepackage{graphicx}
\usepackage{amsmath}
\usepackage{amsthm}
\usepackage{booktabs}
\usepackage{algorithm}
\usepackage[switch]{lineno}

\usepackage{mathtools}
\usepackage{amssymb}
\usepackage{amsmath}
\usepackage{subcaption}
\usepackage{color}
\usepackage{algorithmicx}
\usepackage{algpseudocode}

\theoremstyle{definition}
\newtheorem{lemma}{Lemma}
\newtheorem{definition}{Definition}

\renewcommand\appendix{\par
\setcounter{section}{0}
\setcounter{subsection}{0}
\gdef\thesection{\Alph{section}}}

\newtheorem{theorem}{Theorem}

\title{What Hides behind Unfairness?\\ Exploring Dynamics Fairness in Reinforcement Learning}

\author{
Zhihong Deng$^1$
\and
Jing Jiang$^1$\and
Guodong Long$^1$\And
Chengqi Zhang$^1$\\
\affiliations
$^1$Australian Artificial Intelligence Institute, University of Technology Sydney
\emails
zhi-hong.deng@student.uts.edu.au,
\{jing.jiang,guodong.long,chengqi.zhang\}@uts.edu.au,
}

\begin{document}

\maketitle

\begin{abstract}
In sequential decision-making problems involving sensitive attributes like race and gender, reinforcement learning (RL) agents must carefully consider long-term fairness while maximizing returns. 
Recent works have proposed many different types of fairness notions, but how unfairness arises in RL problems remains unclear. 
In this paper, we address this gap in the literature by investigating the sources of inequality through a causal lens.
We first analyse the causal relationships governing the data generation process and decompose the effect of sensitive attributes on long-term well-being into distinct components.
We then introduce a novel notion called dynamics fairness, which explicitly captures the inequality stemming from environmental dynamics, distinguishing it from those induced by decision-making or inherited from the past. 
This notion requires evaluating the expected changes in the next state and the reward induced by changing the value of the sensitive attribute while holding everything else constant.
To quantitatively evaluate this counterfactual concept, we derive identification formulas that allow us to obtain reliable estimations from data. 
Extensive experiments demonstrate the effectiveness of the proposed techniques in explaining, detecting, and reducing inequality in reinforcement learning. We publicly release code at \textcolor{blue}{\url{https://github.com/familyld/InsightFair}}.
\end{abstract}

%
%
\section{Introduction}


Algorithmic fairness has emerged as an increasingly important research topic in the era of autonomous decision-making systems driven by machine learning~\cite{kasyFairnessEqualityPower2021,mehrabiSurveyBiasFairness2021,pessachAlgorithmicFairness2023}. 
The widespread application of machine learning algorithms in fields like education~\cite{wolffImprovingRetentionPredicting2013,andersonAssessingFairnessGraduation2019}, finance~\cite{dworkFairnessAwareness2012,leeFairnessAwareLoanRecommendation2014}, and law~\cite{angwinMachineBiasThere,tolanWhyMachineLearning2019} carries the risk of unintentionally perpetuating or introducing biases and discrimination against race, gender, or other types of sensitive attributes. 
Despite numerous approaches developed to address algorithmic fairness, many focus on supervised learning or decision-making problems without state transitions~\cite{hardtEqualityOpportunitySupervised2016,zhangCausalFramework2017,liuDelayedImpactFair2018,patilAchievingFairnessStochastic2021}.
However, fairness is not static~\cite{damourFairnessNotStatic2020}.
In reality, the dynamic nature of the world requires intelligent agents to trace the ongoing interaction with the external environment and adjust their behavior based on environmental feedback. This highlights the need to study long-term fairness in sequential decision-making problems.


Recent works have proposed to formulate and investigate long-term fairness within the framework of Markov Decision Processes (MDPs)~\cite{wenAlgorithmsFairnessSequential2021,chiReturnParityMarkov2022,tangTierBalancingDynamic2023}, which makes this topic closely related to reinforcement learning.
Under this framework, a group fairness problem can be formulated as follows: different demographic groups are represented by a set of MDPs that share the same state and action space and run in parallel, and the agent's decisions trigger group-level state transitions and reward assignments.
In this case, long-term fairness regarding the well-being~\cite{pleckoCausalFairnessAnalysis2022} of each group can be measured by the difference in future returns under the agent-specified policy. 
Some other fairness notions have been proposed as well~\cite{liuDelayedImpactFair2018,mouzannarFairDecisionMaking2019,zhangHowFairDecisions2020,tangTierBalancingDynamic2023}, yet the sources of inequality that lead to unfairness remain unclear.
\begin{figure*}[t]
\centering
\includegraphics[width=1.0\textwidth]{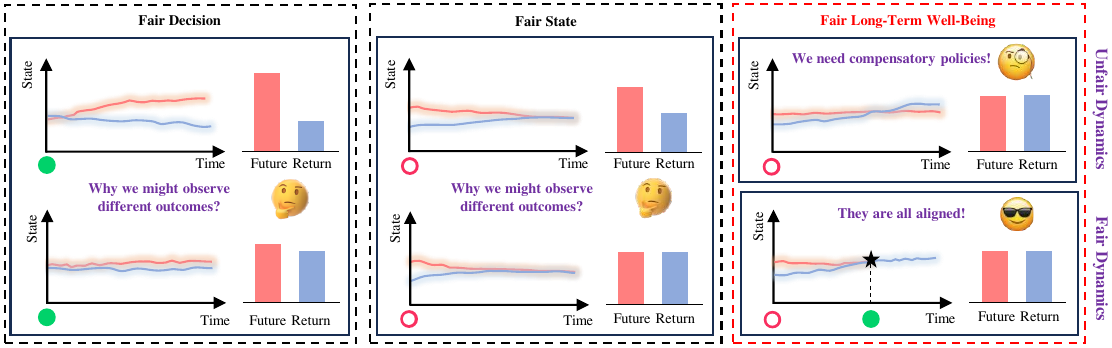}
\caption{Dynamics fairness can help us gain a better understanding of the outcomes of different fairness criteria. This figure shows a sequential decision-making problem with two demographic groups: \textcolor[RGB]{255,125,125}{red} represents the advantaged group and \textcolor[RGB]{143,170,220}{blue} represents the disadvantaged. We present six subfigures, highlighting different fairness criteria (indicated by columns) alongside fair or unfair environmental dynamics (indicated by rows). Each subfigure consists of a line graph showing the evolution of group-wise states over time, accompanied by a histogram illustrating the future returns. The advantaged group may occupy a better initial state (e.g., socio-economic status, qualification profile, expertise level, market competitiveness, etc.), be easier to receive high rewards or reach better states compared to the disadvantaged group. Solid green circles represent consistent decisions for both groups, while empty red circles denote tailored decisions based on sensitive attributes.}
\label{fig1:Example}
\end{figure*}


To highlight the importance of addressing this gap, consider an illustrative example involving two demographic groups, in which the state of each group is represented by a one-dimensional variable, as shown in \figurename~\ref{fig1:Example}. 
The first two columns in this figure show the possible outcomes when two common fairness notions are adopted as criteria. 
The left one corresponds to fair decision, requiring the agent to make consistent decisions for both groups, regardless of sensitive attributes. 
This criterion is popular because people usually expect decision-makers to provide equal resources and opportunities for different groups instead of treating them disparately, but it does not guarantee fair long-term well-being.
The middle column, on the other hand, pursues fair state, urging the agent to improve the socio-economic status or qualification profile of the disadvantaged group, so that different groups will reach similar states in the long run. 
While this notion is very tempting, its implication for long-term well-being is also uncertain, as the disparities would remain if the reward assignment mechanism governing the environment is biased against the disadvantaged group.
Both criteria may lead to different outcomes depending on the environmental dynamics, requiring us to investigate the mechanisms governing the environment, which is overlooked in previous studies.


In this paper, we address the above challenge from a causal perspective. Specifically, we trace the sources of inequality by decomposing the causal effect of sensitive attributes on long-term well-being. 
We formulate the long-term fairness problem via MDP and introduce a novel notion called \emph{dynamics fairness} to capture the fairness of the mechanisms that govern the environment, measuring the expected changes in the next state and the reward induced by changing the identity of a demographic group (the value of a sensitive attribute) while holding everything else constant. 
This notion provides a useful tool for analysing the sources of inequality, distinguishing the inequality introduced by the environmental dynamics from those induced by decision-making or inherited from the past.
Drawing inspiration from research on mediation analysis in the field of causal inference~\cite{pearlCausality2009a,pearlCausalInferenceStatistics2016,pearlDirectIndirectEffects2022}, we propose to characterize dynamics fairness by natural direct effects. 
Since the computation of these effects involves nested counterfactuals that cannot be observed in the factual world, we further derive identification formulas that allow us to reliably estimate these quantities from data. 
As depicted by the right column of \figurename~\ref{fig1:Example}, breaching the fair decision and state criteria is generally inevitable when facing an environment that violates dynamics fairness. 
In this case, to achieve fair long-term well-being, an agent must adopt compensatory policies for the disadvantaged group. 
Conversely, when the environment satisfies dynamics fairness, these criteria can be aligned after eliminating the inequality inherited from the past, as indicated by the star.

In summary, our main contributions are as follows:
\begin{itemize}
\item We introduce a novel causal fairness notion called dynamics fairness. Previous notions were mainly defined in terms of observable disparities in states, decisions, or rewards. To the best of our knowledge, dynamics fairness is the first notion that is defined on the underlying mechanisms governing the environment. Thus, it fills in the missing piece in studying long-term fairness in reinforcement learning.
\item We further derive a method to quantitatively evaluate dynamics fairness through theoretical analysis, which allows this notion to be applied in practical situations. More importantly, this is a general method that does not rely on making parametric assumptions about the environment, such as linearity. 
\item We conducted extensive experiments to verify the theoretical results and demonstrate the effectiveness of the proposed methods in explaining, detecting, and reducing inequality in reinforcement learning problems.
\end{itemize}

\subsection{Preliminaries}

In this paper, we apply causal inference techniques to explore the mechanisms that explain the inequality in reinforcement learning. 
We adopt the framework of structural causal models (SCMs)~\cite{pearlCausality2009a,pearlCausalInferenceStatistics2016}, which provide a mathematical formalization of causality. 
SCMs, denoted as $\mathcal{C} \coloneqq \langle \mathcal{V}, \mathcal{U}, \mathcal{F}, P(u)\rangle$, consist of endogenous (observable) variables $\mathcal{V}$ and background (latent) variables $\mathcal{U}$, linked by structural equations $f_{V_i}$ in $\mathcal{F}$. 
These equations model the causal mechanisms among variables, relating each $V_i \in \mathcal{V}$ to its parents $\text{pa}(V_i)$ and associated exogenous variables $U_{V_i}$, under a probability distribution $P(u)$ over $\mathbf{U}$. 
An SCM induces a causal diagram $\mathcal{G}$ over the nodes $\mathcal{V}$, in which a directed edge $V_i \rightarrow V_j$ exists if $V_i$ is an argument of $f_{V_j}$.

SCMs provide a formal language for articulating counterfactual statements, allowing researchers to examine hypothetical scenarios by evaluating potential outcomes under varied situations. 
Specifically, researchers use $Y_u(x)$ to denote the potential outcome of a variable $Y$ obtained by solving the equations in $\mathcal{F}$ with $X = x$ for an individual $U = u$. 
In causal inference, many research efforts focus on predicting causal effects such as the total effect (TE) $\text{TE}_{x,x'}(Y) \coloneqq \mathbb{E}[Y(x')] - \mathbb{E}[Y(x)]$, where $X$ is the treatment variable and $Y$ is the outcome variable. 
TE measures the difference between the potential outcomes of a variable under different treatments for the whole population (rather than for a specific individual $u$). 
In this paper, we are interested in the TE of the sensitive attribute on the future return, i.e., how a change in identity would affect a demographic group's long-term well-being if all other factors were held constant?~\footnote{While in reality manipulating one's identity is generally infeasible, one can still obtain reliable estimates of the causal effect of interest under appropriate structural assumptions~\cite{pearlCausality2009a}.} 
The following section explores decomposing this effect into distinct components, examining the sources of inequality in RL problems, particularly focusing on the role of environmental dynamics.

\begin{figure}[t]
\centering
\includegraphics[width=1.0\linewidth]{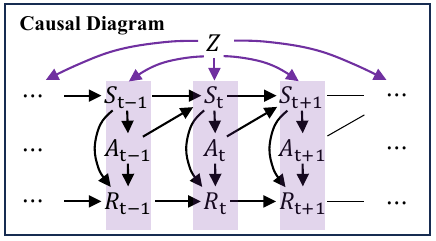} 
\caption{A causal diagram for sequential decision-making problems with a sensitive attribute $Z$. For clarity, the edges from the sensitive attribute to states, decisions, and rewards are represented by the connection between purple arrows and shaded areas.}
\label{fig2:CausalDiagram}
\end{figure}
%
%
\section{Analysing inequality in reinforcement learning}

In this section, we aim to explore and explain the inequality in reinforcement learning through a causal lens. 
We first present the problem formulation, employing causal modeling to analyse environmental dynamics in sequential decision-making scenarios. 
Next, we conduct an in-depth analysis of inequality and investigate its sources through the decomposition of inequality. 
Furthermore, we introduce dynamics fairness and propose an approach for evaluating this counterfactual concept quantitatively. 
Lastly, we present a model-based reinforcement learning algorithm that incorporates dynamics fairness into the planning process.

\subsection{Problem Formulation}
\label{subsec2.1:ProblemFormulation}

\textbf{Problem setup}. 
In reinforcement learning, sequential decision-making problems under uncertainty are often modeled as an MDP $\mathcal{M} = (\mathcal{S}, \mathcal{A}, P, R, \mu, \gamma)$. 
$\mathcal{S}$ and $\mathcal{A}$ denote the state and the action space respectively; 
$P: \mathcal{S} \times \mathcal{A} \rightarrow \Delta_S$ is the transition kernel, where $P(s'|s,a)$ is the probability of transitioning to state $s'$ after taking action $a$ in state $s$; 
$R: \mathcal{S} \times \mathcal{A} \rightarrow \mathbb{R}$ specifies the immediate reward received after taking action $a$ in state $s$; 
$\mu \in \Delta_S$ stands for the initial state distribution. 
At step $t$, the goal of an RL agent is to maximize the future return $G_t = R_{t} + \gamma R_{t+1} + \gamma^2 R_{t+2} + \dots = \sum_{k=0}^{\infty} \gamma^k R_{t+k}$, where $\gamma \in [0,1)$ is the discount factor.
In model-based RL, a common practice is to approximate the real environment using a parametric model $f_{\theta}$, which is employed to generate additional training data for policy learning or predict future transitions and rewards for planning.
To model the environmental dynamics of multiple demographic groups, we consider $N$ MDPs, all sharing the same state and action space, with each MDP representing a distinct demographic group. 
We abstract the group identity as a variable $Z$, referred to as the sensitive attribute. 
Then we can characterize the set of all $N$ MDPs through an augmented dynamics model $f^*_{\theta}$ that takes $Z$ as an additional input. 
This is a common strategy used to address the dynamics generalization problem ~\cite{leeContextawareDynamicsModel2020,guoRelationalInterventionApproach2022},  enabling agents to learn the environment efficiently and accurately.

\noindent\textbf{SCM representation of the dynamics model}. 
Based on the above formulation, we can cast the augmented dynamics model into an SCM $\mathcal{C}$ using auto-regressive uniformization (See Lemma 2 in~\cite{buesingWouldaCouldaShoulda2019}). 
For example, let's consider the conditional distribution $P(R|S,A,Z)$ and denote its cumulative distribution function as $F(R|S,A,Z)$. 
We can construct a new variable $\bar{R} \coloneqq F^{-1}(U_{R}|S,A,Z)$ where $U_{R}$ is a uniform random variable sampled from $[0,1]$. Since the distribution of $\bar{R}$ matches that of $R$, by defining $f_{R} \coloneqq F^{-1}(U_{S'}|S,A,Z)$, we obtain the structural equation for $R$. 
A similar approach can be applied to other variables as well. 
The causal diagram $\mathcal{G}$ induced by the resulting SCM $\mathcal{C}$ is illustrated in \figurename~\ref{fig2:CausalDiagram}. 
Note that the impact of the sensitive attribute may cover various facets of the system, including state transitions, decision-making, and reward assignments. 
Therefore, inequality may stem from multiple sources following different causal paths in $\mathcal{G}$.

\begin{figure*}[t]
\centering
\includegraphics[width=0.99\linewidth]{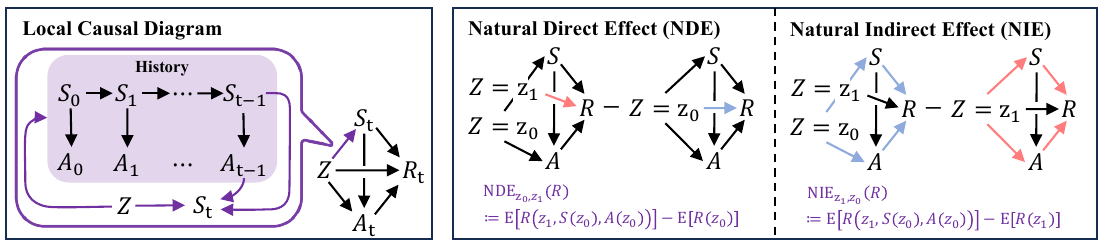} 
\caption{(Left) A local causal diagram in which the directed edge $Z \rightarrow S_t$ carries the direct effect of $Z$ on $S_t$ and the indirect effect mediated by past states and actions (the history). 
We can combine them into one when focusing on analysing the effect of $Z$ on $R_t$ (or $S_{t+1}$). 
(Right) A graphical representation of natural direct and indirect effects, where the contrast between the two quantities is highlighted in \textcolor[RGB]{255,125,125}{red} and \textcolor[RGB]{143,170,220}{blue}.}
\label{fig3:LocalCausalDiagramAndCausalEffects}
\end{figure*}
\subsection{Decomposition of Inequality}

In this work, we focus on the inequality in long-term well-being, a causal quantity that is captured by the difference in future return $G_t$ when setting the sensitive attribute $Z=z_1$ compared to $Z=z_0$. Using the causal language, we can now formally define this quantity as follows. 
\begin{definition}[Well-being Gap]
At step $t$, the well-being gap between the two demographic groups distinguished by the sensitive attribute $Z$ is defined as:
$$\text{TE}_{z_0,z_1}(G_t) \coloneqq \mathbb{E}[G_t(z_1)] - \mathbb{E}[G_t(z_0)].$$
\end{definition}
This quantity measures the total effect of the sensitive attribute $Z$ on the future return $G_t$. Since $G_t$ is a linear combination of the reward signals, the well-being gap satisfies the following lemma:
\begin{lemma}
\label{lem1:GapDecomposition}
The well-being gap $\text{TE}_{z_0,z_1}(G_t)$ can be decomposed into the sum of reward gaps at each time step:
$$\text{TE}_{z_0,z_1}(G_t) = \sum_{k=0}^{\infty} \gamma^k \text{TE}_{z_0,z_1}(R_{t+k}),$$
where $\text{TE}_{z_0,z_1}(R_{t}) \coloneqq \mathbb{E}[R_t(z_1)] - \mathbb{E}[R_t(z_0)]$ is the reward gap at step $t$.
\end{lemma}
Lemma~\ref{lem1:GapDecomposition} indicates that we can focus on the reward gap at each time step to examine the sources of inequality in long-term well-being. 
In order to analyse the reward gap $\text{TE}_{z_0,z_1}(R_{t})$ at step $t$, we present the local causal diagram in \figurename~\ref{fig3:LocalCausalDiagramAndCausalEffects} (Left). 
From the diagram, we can see that $\text{TE}_{z_0,z_1}(R_{t})$ is affected by the sensitive attribute $Z$ through two types of causal paths: the direct path $Z \rightarrow R_t$ and the indirect paths mediated by the current state $S_t$ or the decision $A_t$. 
Notably, $Z \rightarrow S_t$ carries the inequality inherited from historical factors as well as the one directly caused by $Z$.  

Next, we introduce two counterfactual quantities that are useful for a refined decomposition of the reward gap.  
We omit the subscript $t$ when the context is clear.
\begin{definition}[Natural Direct and Indirect Effect]
The natural direct and indirect effects of sensitive attribute $Z$ on reward $R$ are defined, respectively, as:
$$\text{NDE}_{z_0,z_1}(R) \coloneqq \mathbb{E}[R(z_1, S(z_0), A(z_0))] - \mathbb{E}[R(z_0)].$$
$$\text{NIE}_{z_1,z_0}(R) \coloneqq \mathbb{E}[R(z_1, S(z_0), A(z_0))] - \mathbb{E}[R(z_1)].$$
\end{definition}
\figurename~\ref{fig3:LocalCausalDiagramAndCausalEffects} (Right) illustrates these two quantities, both involving nested counterfactuals. 
For instance, in $\text{NDE}_{z_0,z_1}(R)$, the first factor involves computing the expected reward when $S$ and $A$ are set to the values they would have naturally taken under $Z=z_0$ while $Z$ is set to $z_1$. 
Perhaps surprisingly, despite being not realizable in the factual world, these quantities are of great interest and are widely used in many practical applications. 
In discrimination cases, by keeping the state $S$ and decision $A$ to the level they would have naturally attained under $Z=z_0$, $\text{NDE}_{z_0,z_1}(R)$ measures the advantage that the individuals in the disadvantaged group $Z=z_0$ would have gained if there were no direct discrimination within the reward assignment process. Conversely, $\text{NIE}_{z_1,z_0}(R)$ measures the advantage that the individuals in the disadvantaged group $Z=z_0$ lose due to indirect discrimination.

Based on Lemma~\ref{lem1:GapDecomposition} and the above definitions, we can now present a quantitative decomposition of the well-being gap:
\begin{theorem}[Causal Decomposition of Well-being Gap]
\label{thm1:Decomposition}
The well-being gap $\text{TE}_{z_0,z_1}(G_t)$ can be decomposed as follows:
$$\text{TE}_{z_0,z_1}(G_t) = \sum_{k=0}^{\infty} \gamma^k \left(\text{NDE}_{z_0,z_1}(R) - \text{NIE}_{z_1,z_0}(R)\right).$$
\end{theorem}
Theorem~\ref{thm1:Decomposition} offers a general decomposition of the well-being gap, quantitatively explaining how it is accumulated over time,  without assuming any specific functional form of the structural equations in $\mathcal{C}$. The following lemma further reveals the qualitative relationship between the natural direct and indirect effects and the causal paths.
\begin{lemma}
\label{lem2:Relationship}
Consider a set of MDPs described by the augmented dynamics model $f^*_{\theta}$, along with the induced causal diagram $\mathcal{G}$, the following statements hold: If $\text{NDE}_{z_0,z_1}(R) \neq 0$, then there exists a direct path from $Z$ to $R$; Similarly, if $\text{NIE}_{z_1,z_0}(R) \neq 0$, then there exist indirect paths connecting $Z$ and $R$. 
\end{lemma}
That is to say, the natural direct and indirect effects provide hints for structural information about the causal diagram, thus helping us identify the causal paths that are responsible for the gap. 
Summing up, the analysis presented herein systematically examines the sources of inequality through a causal lens.
By introducing the natural direct and indirect effects, we provide both quantitative and qualitative explanations of the well-being gap, offering insights into the intricacies of inequality in reinforcement learning problems.

\subsection{Dynamics Fairness}

Beyond merely explaining the sources of inequality, in practice, we also seek clarity regarding the responsibility for such inequality.
Theorem~\ref{thm1:Decomposition} and Lemma~\ref{lem2:Relationship} demonstrate that the inequality accumulated at each step can be attributed to two types of causal paths: 
the direct path ($Z \rightarrow R$) and the indirect paths ($Z \rightarrow S \rightarrow R$, $Z \rightarrow A \rightarrow R$, and $Z \rightarrow S \rightarrow A \rightarrow R$). 
However, the inequality introduced by the environmental dynamics and the decisions is still coupled together.
In other words, $\text{NIE}_{z_0,z_1}(R) \neq 0$ can not tell us how to determine responsibility between the two.
Before diving into further details, we first formally define the notion of dynamics fairness.
\begin{definition}[Dynamics Fairness]
\label{def3:DynamicsFairness}
The dynamics is fair regarding the sensitive attribute $Z$ if there are no direct paths from $Z$ to either the reward $R$ or the next state$S'$ in the causal diagram $\mathcal{G}$ induced by the augmented dynamics model $f^*_{\theta}$.
\end{definition}
Definition~\ref{def3:DynamicsFairness} provides a structural condition for the environment to satisfy dynamics fairness. 
Banning the direct paths from the sensitive attribute to the reward and the next state ensures inequality is not introduced by the environmental dynamics. 
As a result, there are only two possible sources of inequality: historical inequality, i.e., the disparity in the current state of different demographic groups, and the inequality introduced by the decisions, i.e., disparate treatment in the decision-making process. 
Specifically, the historical inequality can be estimated by the observed disparity in the current state $\mathbb{E}[S|Z=z_1] - \mathbb{E}[S|Z=z_0]$, which is easy to measure. 
If the environment satisfies dynamics fairness and there is no historical inequality, then the well-being gap is completely attributed to the decisions. 
In this way, we can solve the problem of attributing responsibility for inequality.
Building upon Lemma~\ref{lem2:Relationship}, we establish the following theorem:
\begin{theorem}[Criterion for Dynamics Fairness Violation]
\label{thm2:Violation}
The environment is considered to violate dynamics fairness if either $\text{NDE}{z_0,z_1}(R) \neq 0$ or $\text{NDE}{z_0,z_1}(S') \neq 0$ holds.
\end{theorem}
Theorem~\ref{thm2:Violation} offers a quantitative criterion for dynamics fairness. 
By checking the natural direct effects of $Z$ on reward $R$ and next state $S'$, we can determine whether the mechanisms governing the environmental dynamics would introduce inequality during the sequential decision-making process, thus contributing to the well-being gap in the long run. 
More importantly, it also provides mathematical justification for compensatory policies. 
If the environment violates dynamics fairness, then disparate treatment is necessary to compensate the disadvantaged group when aiming to achieve equality in long-term well-being. 
On the other hand, if the environment is fair, then disparate treatment should be limited to a minimum level, not only for the sake of fair decision-making but also to avoid widening the well-being gap.

As mentioned earlier, the computation of $\text{NDE}_{z_0,z_1}(\cdot)$ involves nested counterfactuals, which require data that we cannot obtain in the factual world. 
Fortunately, under the structural assumptions implied by the augmented dynamics model, we can derive identification equations for these effects.
\begin{theorem}[Identification of Dynamics Fairness]
\label{thm3:Identification}
Under the structural assumptions implied by the augmented dynamics model depicted by the local causal diagram, the natural direct effects of $Z$ on $R$ and $S'$ can be identified as follows:
\begin{align*}
&\text{NDE}_{z_0,z_1}(R) = \sum_{s,a} \big (\mathbb{E}[R | Z=z_1, S=s, A=a]\\ - &\mathbb{E}[R | Z=z_0, S=s, A=a] \big ) P(S=s, A=a | Z=z_0).
\end{align*}
\begin{align*}
&\text{NDE}_{z_0,z_1}(S') = \sum_{s,a} \big (\mathbb{E}[S' | Z=z_1, S=s, A=a]\\ - &\mathbb{E}[S' | Z=z_0, S=s, A=a] \big ) P(S=s, A=a | Z=z_0).
\end{align*}
\end{theorem}
In the above equations, all quantities on the right-hand side are expressed using conditional expectations or probabilities rather than counterfactuals, and they can be estimated directly from data using standard statistical methods. 
As a result, Theorem~\ref{thm3:Identification} provides a practical way to determine whether the environment satisfies dynamics fairness. Proofs are included in Appendix~\ref{app1:Proofs} in the supplementary material.

\subsection{Algorithm for Achieving Long-Term Fairness}

So far, we have examined the sources of inequality and the associated responsibilities, with a particular focus on the influence of environmental dynamics.
The presented results not only demystify the intricate relationships behind unfairness in RL problems but also provide valuable guidance for designing RL algorithms to tackle inequality.
Based on the above analyses, we propose a model-based RL algorithm named \emph{InsightFair} that incorporates dynamics fairness into the planning process.
Specifically, the agent iteratively collects environmental data and trains an ensemble augmented dynamics model $f^{*}_{\theta}$, which is later used to simulate the environment and evaluate the action sequences for planning.
At each epoch, the agent first evaluates dynamics fairness following Theorem~\ref{thm3:Identification} using $f^{*}_{\theta}$ and the collected data.
If the environment violates dynamics fairness, then disparate treatment is considered necessary in the planning process. 
Otherwise, the agent samples consistent action sequences for all groups when the observed disparity in the current state approaches zero.
The optimization objective is regularized by a constraint on the estimated well-being gap to induce a fair future. 
The detailed algorithm is presented in Appendix~\ref{app2:Algorithm}.

%
%
\section{Experiments}
\label{sec4:Experiments}
In this section, we empirically validate the theoretical results of our work through extensive experiments. Specifically, the results presented here aim to answer the following questions: (1) Does the proposed decomposition offer an accurate quantitative explanation for inequality? (2) Can the proposed identification formula reliably detect cases where the environment violates dynamics fairness? (3) Does the proposed algorithm, InsightFair, work effectively with the aid of dynamics fairness? 
To be self-contained, we include additional details and results in Appendix~\ref{app3：Experiments} in the supplementary material.

\subsection{Explaining Inequality}
\label{subsec3.1:ExplainingInequality}

To answer the first question, we consider a simple non-linear model for the reward function, i.e., $R = 1$ if $w_0 + w_1 z + w_2 s + w_3 a + u_{R} \geq 0$ and $R = 0$ otherwise, where $w_0, w_1, w_2, w_3$ are the model parameters and $U_{R} \sim \text{logistic}(0,1)$ is the background variable. Since the reward function is fixed within an episode and the return is a linear combination of the rewards, we can examine the well-being gap by analysing the decomposition of the reward gap $\text{TE}_{z_0,z_1}(R)$. 

\begin{figure}[t]
\centering
\includegraphics[width=1.0\linewidth]{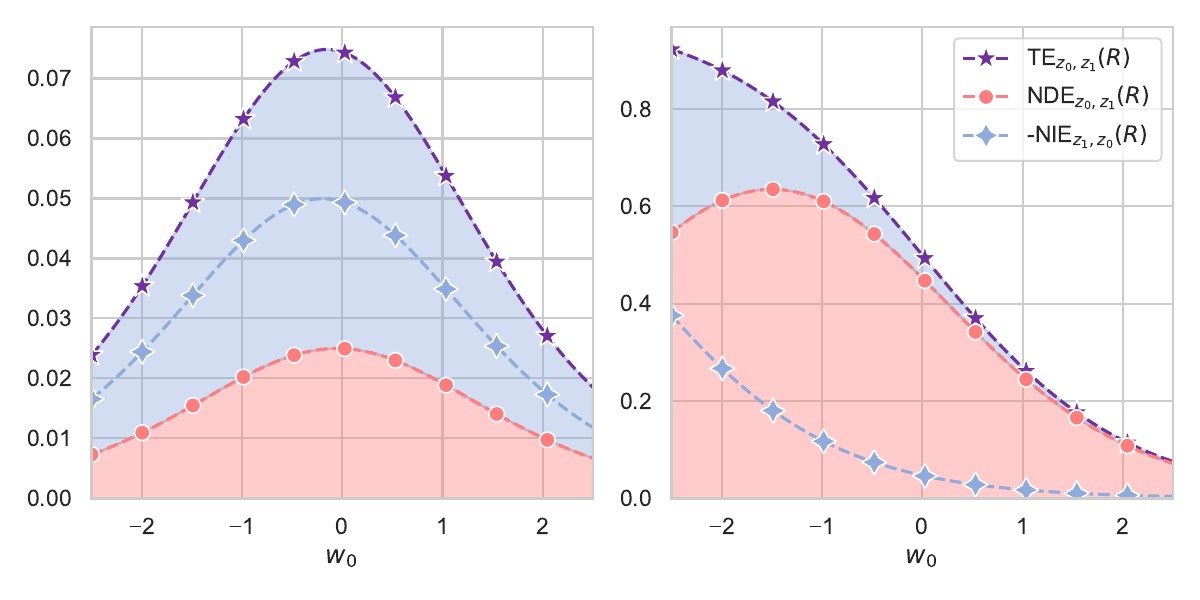} 
\caption{Visualization of total effect, natural direct effect, and natural indirect effect under different model parameters. The \textcolor[RGB]{112,48,160}{purple} curve represents $\text{TE}_{z_0,z_1}(R)$, while the \textcolor[RGB]{255,125,125}{red} and \textcolor[RGB]{143,170,220}{blue} curves depict $\text{NDE}_{z_0,z_1}(R)$ and $-\text{NIE}_{z_1,z_0}(R)$, respectively. The two figures illustrate scenarios where either the indirect effect dominates (top) or the direct effect dominates (bottom), while the shaded areas show that the sum of $\text{NDE}_{z_0,z_1}(R)$ and $-\text{NIE}_{z_1,z_0}(R)$ matches the total effect in both scenarios, as guaranteed by Theorem~\ref{thm1:Decomposition}.}
\label{fig4:Explain}
\end{figure}

\figurename~\ref{fig4:Explain} illustrates the comparison between the total effect $\text{TE}_{z_0,z_1}(R)$ and the natural direct and indirect effects $\text{NDE}_{z_0,z_1}(R)$ and $\text{NIE}_{z_1,z_0}(R)$, respectively. The values are computed analytically under two sets of model parameters $w_1=w_2=w_3=0.1$ and $w_1=w_2=w_3=3.0$ along with $w_0$ varying in the range $[-2.5,2.5]$. First, we can see that the sum of $\text{NDE}_{z_0,z_1}(R)$ and $-\text{NIE}_{z_1,z_0}(R)$ accurately explains the reward gap $\text{TE}_{z_0,z_1}(R)$ in both settings. Second, by varying $w_0$, the proportion of the direct and indirect effects changes accordingly. In the right figure, the dominance of the direct effect suggests that the inequality is mainly introduced by the environmental dynamics instead of other factors.

\begin{figure}[t]
\centering
\includegraphics[width=1.0\linewidth]{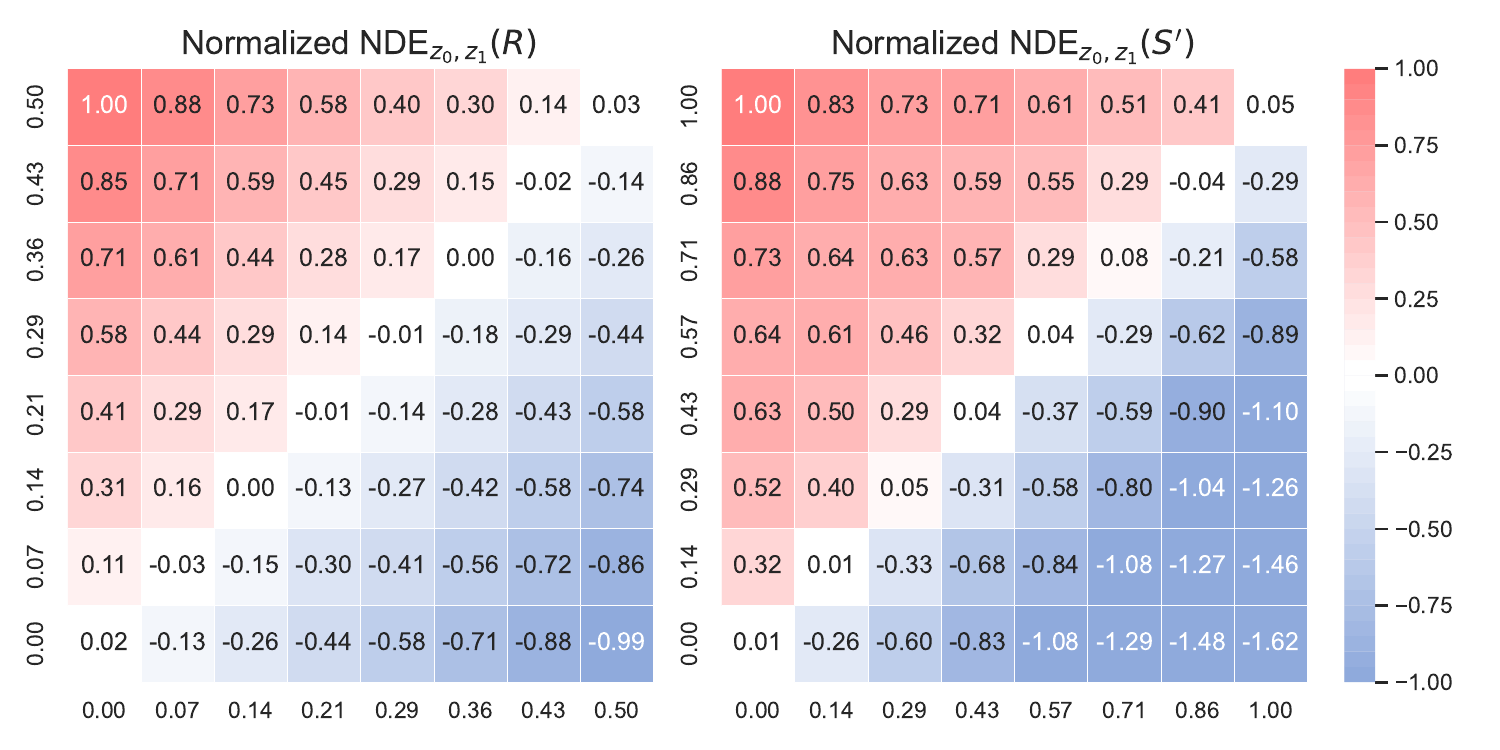} 
\caption{Results of evaluating dynamics fairness using the proposed identification formula. Each tile represents the estimated natural direct effects, $\text{NDE}{z_0,z_1}(R)$ or $\text{NDE}{z_0,z_1}(S')$, under specific parameter configurations indicated by the row and column. Lighter colors signify values closer to zero (satisfying dynamics fairness), while darker colors represent larger absolute values (violating dynamics fairness). \textcolor[RGB]{255,125,125}{Red} denotes the second demographic group is advantaged, while \textcolor[RGB]{143,170,220}{blue} denotes disadvantaged.}
\label{fig5:Detect}
\vskip -0.2in
\end{figure}
\begin{figure*}[t]
\centering
\begin{subfigure}{0.42\linewidth}
\includegraphics[width=0.99\linewidth]{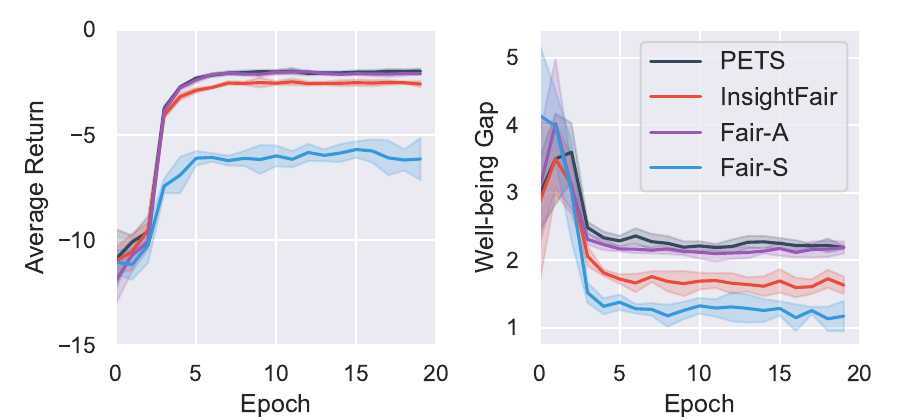} 
\end{subfigure}
\begin{subfigure}{0.28\linewidth}
\includegraphics[width=0.99\linewidth]{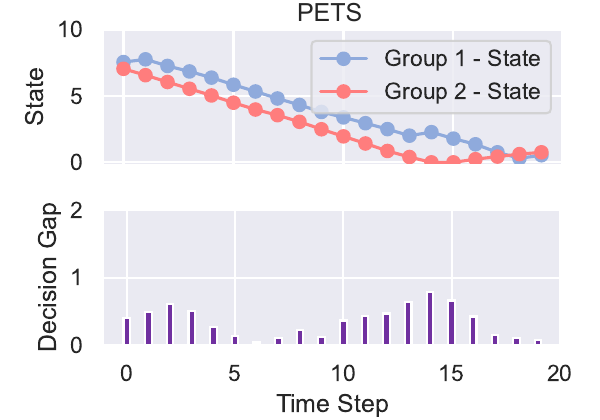}
\end{subfigure}
\begin{subfigure}{0.28\linewidth}
\includegraphics[width=0.99\linewidth]{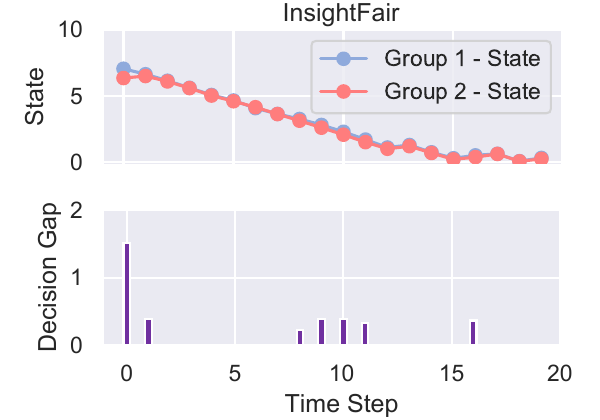}
\end{subfigure}
\caption{(Left) Learning curves depicting return and the well-being gap. (Right) Changes in state and decision gaps within an episode.}
\label{fig6:Comparison}
\end{figure*}
\subsection{Detecting Violation of Dynamics Fairness}
\label{subsec3.2:DetectingViolation}

In this part, we empirically verify the effectiveness of the proposed identification formula by customizing the ML-fairness-gym suite~\cite{damourFairnessNotStatic2020} for group fairness problems and code the proposed algorithms using Omnisafe~\cite{omnisafe2023}. 
We use the Allocation environment that contains two demographic groups. At each step, some incidents like rat infestations occur in both groups, depending on their current state. The agent must allocate resources effectively to minimize both missed incidents and resources allocated. Groups do not interact with each other. By manipulating the reward function and the transition dynamics, we can generate different environments to simulate scenarios where the environment satisfies or violates dynamics fairness. 

We conduct two sets of experiments to evaluate $\text{NDE}_{z_0,z_1}(R)$ and $\text{NDE}_{z_0,z_1}(S')$, respectively. In the first set, we vary the reward function while keeping the transition dynamics fixed. Two parameters $\beta_0$ and $\beta_1$ are used to control the relative advantage of the second demographic group. When $\beta_1 > \beta_0$, the second group is advantaged, and vice versa. The environment is considered fair when $\beta_1 = \beta_2$, i.e., the relative advantage reduces to zero. The second set of experiments is conducted similarly, except that we fix the reward function and vary the parameters controlling the transition dynamics to determine the relative advantage. For both sets, we conduct a total of 64 experiments by varying the parameters in a fixed range to generate a series of environments and estimating the natural direct effects using the proposed identification formula. The results are shown in \figurename~\ref{fig5:Detect}. From this figure, it is clear that the natural direct effects approach zero when the environment satisfies dynamics fairness (as colored in white on the diagonal). On the other hand, when the environment violates dynamics fairness, the natural direct effects vary with the magnitude of the relative advantage, as indicated by the darker colors in the heatmaps. These results empirically support that the proposed approach can reliably detect violations of dynamics fairness.

\subsection{Promoting Long-Term Fairness}
\label{subsec3.3:PromotingLongtermFairness}

To answer the third question, we evaluate the performance of the proposed algorithm by comparing it with the base algorithm, PETS~\cite{chuaDeepReinforcementLearning2018}, which is a popular model-based RL algorithm but does not consider fairness. In addition, we consider two variants of PETS. The first variant, Fair-A, requires the agent to satisfy the fair decision criterion, sampling consistent action sequences for all groups. The second variant, Fair-S, incorporates a constraint regarding state disparities into the optimization objective, aiming to achieve equality in the state for all groups. Existing studies~\cite{wenConstrainedCrossEntropyMethod2018,liuConstrainedModelbasedReinforcement2021} have shown that model-based algorithms can effectively satisfy constraints while maintaining good task performance. Note that our focus in this paper is orthogonal to these works. The experiments presented herein aim to examine the effectiveness of the proposed fairness notion when incorporated into a model-based RL algorithm, particularly in promoting fair long-term well-being when the environment violates dynamics fairness and in aligning different criteria in a fair environment.

The learning curves in \figurename~\ref{fig6:Comparison} (Left) illustrate the average return and the well-being gap for all algorithms in the Allocation environment. The results are averaged over five runs with different random seeds. InsightFair exhibits a slightly lower task performance compared to PETS and Fair-A, while considerably reducing the well-being gap, indicating its effectiveness in promoting long-term fairness when the environment violates dynamics fairness. Fair-S also reduces the well-being gap but at the cost of task performance, as it sacrifices the utility of the advantaged group to achieve equality in the state. In the next experiment shown in \figurename~\ref{fig6:Comparison} (Right), we consider a fair environment where the relative advantage is zero but the advantaged group begins with a better initial state. We visualize the changes in state and decision gap (the absolute difference in actions) between the two groups. InsightFair makes consistent decisions for both groups when state disparities approach zero, resulting in zero decision gap in most steps, while PETS demonstrates disparate treatment throughout the episode. These results suggest that InsightFair effectively aligns the fair state, action, and well-being criteria in a fair environment.


%
%
\section{Related Work}

\textbf{Long-term fairness.}
Consideration of fairness in machine learning systems has evolved beyond static settings to a dynamic perspective~\cite{damourFairnessNotStatic2020}, i.e., autonomous agents may interact with the environment over time and have a long-term impact on fairness. 
Notably, growing evidence~\cite{liuDelayedImpactFair2018,zhangHowFairDecisions2020} support concerns about static fairness measures, as they may jeopardize long-term fairness. 
In recent years, there has been a considerable amount of work devoted to promoting long-term fairness~\cite{mouzannarFairDecisionMaking2019,wenAlgorithmsFairnessSequential2021,puranikDynamicDecisionMakingFramework2022,chiReturnParityMarkov2022,tanRISERobustIndividualized2022,yuPolicyOptimizationAdvantage2022,xuEqualLongtermBenefit2023}.
\cite{wenAlgorithmsFairnessSequential2021} investigated the long-term fairness problem under scenarios in which individuals receive decisions repeatedly. 
Specifically, they define fairness as the gap between the expected cumulative rewards of individuals in different groups. 
The proposed algorithm relies on a known model and can be reduced to solving a linear programming problem in the finite-state MDPs setting. 
\cite{chiReturnParityMarkov2022} adopt a similar definition of fairness. 
Their algorithm employs integral probability metrics to align the state visitation distributions among groups, thereby mitigating return disparity. 
More recently, \cite{yuPolicyOptimizationAdvantage2022} proposed a policy optimization algorithm that incorporates fairness constraints into the advantage function. 
This enables an agent to efficiently explore the environment and learn a policy that strikes the balance between utility and fairness.
However, \cite{xuEqualLongtermBenefit2023} noted that assuming the fairness metric as part of the state is generally impractical because long-term fairness necessitates the agent to focus on the future rather than the past. 
Their work studied supply-demand MDPs, a variant of MDPs that contain two time-varying variables, supply and demand, for each group depending on states and actions. 
The authors suggested using the ratio between the cumulative group supply and demand to measure long-term well-being and proposed a policy optimization algorithm that can effectively reduce the discrepancy in this ratio.
Additionally, multi-objective RL algorithms~\cite{reymond2022pareto} have shown potenial in handling the complex trade-offs inherent in balancing various fairness criteria.
In this paper, we systematically examine long-term fairness for demographic groups characterized by a set of different MDPs.  
Our work features causal modeling of the agent-environment interaction and aims to identify the sources of inequality, thereby providing a new perspective on studying the long-term fairness problem.

\noindent\textbf{Causal fairness analysis.} 
Causal fairness analysis, an emerging trend in fairness research, leverages techniques and theories from the causal inference community to advance the exploration of fairness problems in machine learning~\cite{creagerCausalModelingFairness2020,pleckoCausalFairnessAnalysis2022,dengCausalReinforcementLearning2023a}. 
This methodology not only rigorously defines fairness but also provides effective and reliable solutions to ensure it."
For example,~\cite{kusnerCounterfactualFairness2017} proposed counterfactual fairness for fair prediction problems, requiring the distribution over possible predictions of an individual to remain unchanged in a world where its sensitive attributes had been changed. 
In fact, counterfactuals provide a versatile framework for formulating various fairness notions of different granularity. \cite{zhangFairnessDecisionMakingCausal2018} introduced different types of counterfactual effects for the fair decision problem. 
These effects offer fairness notions at the group level, helping researchers interpret various discrimination mechanisms. 
Furthermore, \cite{nabiLearningOptimalFair2019} and~\cite{wuPCFairnessUnifiedFramework2019} demonstrated how causal graphs can assist researchers in defining fine-grained fairness associated with specific causal paths. 
For instance, using sensitive attributes or their proxies is deemed reasonable in some cases due to business necessity~\cite{pleckoCausalFairnessAnalysis2022} but may lead to discrimination in other cases. 
Recent research has extended the study of causal fairness to contextual bandits and outcome control problems~\cite{huangAchievingUserSideFairness2022,pleckoCausalFairnessOutcome2023a}.
Unlike previous studies, our paper delves into the fairness problem in the sequential decision-making setting, introducing a novel causal fairness notion, dynamics fairness, which captures the fairness of the mechanisms governing the environment.

%
%
\section{Conclusion}

In this paper, we systematically studied inequality in RL problems through a causal lens, aiming to demystify the intricate relationships behind unfairness. We introduced a causal model to formalize the agent-environment interaction and provided a quantitative decomposition of the well-being gap, which explains how inequality is accumulated over time. Furthermore, we proposed dynamics fairness, a novel causal fairness notion that, for the first time, captures the fairness of the underlying mechanisms governing the environment. To measure this counterfactual quantity, we proposed identification formulas that do not rely on specific functional forms or a known model. Extensive experiments were conducted to verify the theoretical results and demonstrate the effectiveness of the proposed approaches. In the future, we plan to extend our work to more complicated scenarios, such as partially observable and non-stationary environments.

\clearpage

\bibliographystyle{named}
\bibliography{ijcai24}

\clearpage

\appendix


\section{Proofs}
\label{app1:Proofs}
\begin{proof}[\textbf{Proof of Lemma~\ref{lem1:GapDecomposition}}]
The well-being gap $\text{TE}_{z_0,z_1}(G_t)$ is defined as the difference between two expectations:
$$\text{TE}_{z_0,z_1}(G_t) \coloneqq \mathbb{E}[G_t(z_1)] - \mathbb{E}[G_t(z_0)],$$
in which $\mathbb{E}[G_t(z)]$ denote the expected value of potential outcomes of return $G_t$ when the sensitive attribute $Z$ is set to $z$. We can expand $G_t$ as the sum of discounted rewards. Since the expected value operator $\mathbb{E}[\cdot]$ is linear, we have:
\begin{align*}
\mathbb{E}[G_t(z)] &= \mathbb{E}[R_t(z) + \gamma R_{t+1}(z) + \gamma^2 R_{t+2}(z) + \cdots]\\
&= \mathbb{E}[R_t(z)] + \gamma \mathbb{E}[R_{t+1}(z)] + \gamma^2 \mathbb{E}[R_{t+2}(z)] + \cdots\\
&= \sum_{k=0}^{\infty} \gamma^k \mathbb{E}[R_{t+k}(z)].
\end{align*}
As a result, the well-being gap can be decomposed as the discounted sum of reward gaps at each time step:
\begin{align*}
\text{TE}_{z_0,z_1}(G_t) &= \sum_{k=0}^{\infty} \gamma^k \mathbb{E}[R_{t+k}(z_1)] - \sum_{k=0}^{\infty} \gamma^k \mathbb{E}[R_{t+k}(z_0)]\\ 
&= \left(\mathbb{E}[R_t(z_1)] - \mathbb{E}[R_t(z_0)]\right)\\
&+ \gamma \left(\mathbb{E}[R_{t+1}(z_1)] - \gamma \mathbb{E}[R_{t+1}(z_0)]\right) + \cdots\\
&= \sum_{k=0}^{\infty} \gamma^k \text{TE}_{z_0,z_1}(R_{t+k}).
\end{align*}
\end{proof}

\begin{proof}[\textbf{Proof of Theorem~\ref{thm1:Decomposition}}]
Based on Lemma~\ref{lem1:GapDecomposition}, to prove this theorem, it suffices to show that the reward gap $\text{TE}_{z_0,z_1}(R_{t})$ can be decomposed as the sum of $\text{NDE}_{z_0,z_1}(R)$ and $- \text{NIE}_{z_1,z_0}(R)$, which can be derived straightforwardly from the definitions of the natural direct and indirect effects:
\begin{align*}
\text{TE}_{z_0,z_1}(R_{t}) &= \mathbb{E}[R_t(z_1)] - \mathbb{E}[R_t(z_0)]\\
&= (\mathbb{E}[R_t(z_1, S(z_0), A(z_0))] - \mathbb{E}[R_t(z_0)])\\
&+ (\mathbb{E}[R_t(z_1)] - \mathbb{E}[R_t(z_1, S(z_0), A(z_0))])\\
&= \text{NDE}_{z_0,z_1}(R_t) - \text{NIE}_{z_1,z_0}(R_t).
\end{align*}
Since the reward function is time-invariant, we have:
$$\text{TE}_{z_0,z_1}(G_t) = \sum_{k=0}^{\infty} \gamma^k \left(\text{NDE}_{z_0,z_1}(R) - \text{NIE}_{z_1,z_0}(R)\right).$$
\end{proof}

\begin{proof}[\textbf{Proof of Lemma~\ref{lem2:Relationship}}]
Consider the contrapositive of the original statement: If there is no direct causal path from $Z$ to $R$ and $S'$ in a causal diagram $\mathcal{G}$ induced by the augmented dynamics model $f^*_{\theta}$, then $\text{NDE}_{z_0,z_1}(R) = 0$ and $\text{NDE}_{z_0,z_1}(S') = 0$; Similarly, if there are no indirect causal paths from $Z$ to $R$, then $\text{NIE}_{z_1,z_0}(R) = 0$. We can divide the proof of this new statement into two parts.

First, following the consistency rule of counterfactuals~\cite[Corollary 7.3.2]{pearlCausality2009a}, we have $R(z_0) = R\left(z_0, S(z_0), A(z_0)\right)$. Since there is no direct path from $Z$ to $R$, we have $R(z, s, a) = R(s, a)$ for all $z, s, a$ according to the exclusion restriction rule of structure-based counterfactuals~\cite[Section~7.3.2]{pearlCausality2009a}. As a result, we can now rewrite the natural indirect effect of $Z$ on $R$ as follows:
\begin{align*}
\text{NDE}_{z_0,z_1}(R) &= \mathbb{E}[R(z_1, S(z_0), A(z_0))] - \mathbb{E}[R(z_0)]\\
&= \mathbb{E}[R(z_1, S(z_0), A(z_0))] - \mathbb{E}[R(z_0, S(z_0), A(z_0))]\\
&= \mathbb{E}[R(S(z_0), A(z_0)) - R(S(z_0), A(z_0))] = 0.
\end{align*}

Similarly, when there are no indirect paths from $Z$ to $R$, or more specifically, when there is no path from $Z$ to $R$ that passes through $S$ or $A$ in $\mathcal{G}$, we have $R(z, s, a) = R(z)$ for all $z, s, a$. As a result, the natural direct effect of $Z$ on $R$ can be rewritten as follows:
\begin{align*}
\text{NIE}_{z_1,z_0}(R) &= \mathbb{E}[R(z_1)] - \mathbb{E}[R(z_1, S(z_0), A(z_0))]\\
&= \mathbb{E}[R(z_1)] - \mathbb{E}[R(z_1)] = 0.
\end{align*}

Summing up, we show that the contrapositive statement is true, and by equivalence, the original statement is proved.
\end{proof}

\begin{proof}[\textbf{Proof of Theorem~\ref{thm2:Violation}}]
Definition~\ref{def3:DynamicsFairness} states that, to satisfy dynamics fairness, the sensitive attribute $Z$ should not causally influence the reward $R$ or the next state $S'$ through direct causal paths in the causal diagram $\mathcal{G}$ induced by the augmented dynamics model $f^*_{\theta}$. Lemma~\ref{lem2:Relationship} shows that $\text{NDE}_{z_0,z_1}(R)$ can be used to detect the direct causal paths from $Z$ to $R$. We can prove that $Z \rightarrow S'$ exists if $\text{NDE}_{z_0,z_1}(S') \neq 0$ in a similar way. As a result, if either $\text{NDE}_{z_0,z_1}(R) \neq 0$ or $\text{NDE}_{z_0,z_1}(S') \neq 0$ holds, then the environment violates dynamics fairness. In other words, we can detect the violation of dynamics fairness by checking the natural direct effects of $Z$ on $R$ and $S'$.
\end{proof}

\begin{proof}[\textbf{Proof of Theorem~\ref{thm3:Identification}}]
Here we focus on the identification of $\text{NDE}_{z_0,z_1}(R)$, as the identification of $\text{NDE}_{z_0,z_1}(S')$ can be proved similarly. We first rewrite the natural direct effect of $Z$ on $R$ as follows:
\begin{align*}
&\text{NDE}_{z_0,z_1}(R) = \mathbb{E}[R(z_1, S(z_0), A(z_0))] - \mathbb{E}[R(z_0)]\\
= &\sum_{s,a} \big (\mathbb{E}[R(z_1, s, a) \mid S(z_0)=s, A(z_0)=a]\\
- &\mathbb{E}[R(z_0, s, a) \mid S(z_0)=s, A(z_0)=a] \big ) P(S(z_0)=s, A(z_0)=a)\\
= &\sum_{s,a} \left(\mathbb{E}[R(z_1, s, a)] - \mathbb{E}[R(z_0, s, a)]\right) P(S(z_0)=s, A(z_0)=a).
\end{align*}
The last step holds because there are no paths between $R$ and $Z, S, A$ in $\mathcal{G}$ containing only background variables. In this case, we have $R(z,w) \perp\!\!\!\perp W(z')$ for any $z, z', w$ according to the independence restriction rule of structure-based counterfactuals~\cite[Section~7.3.2]{pearlCausality2009a}. The resulting equality is also known as the experimental identification of natural direct effects~\cite{pearlDirectIndirectEffects2022}. Since none of the quantities on the right-hand side involve nested counterfactuals, it can be estimated using experimental data. 

Furthermore, based on the rule 2 of do-calculus~\cite[Section~3.4.2]{pearlCausality2009a}, $\mathbb{E}[R(z, s, a)] = \mathbb{E}[R | z, s, a]$ holds for all $z, s, a$ and $P(S(z_0)=s, A(z_0)=a) = P(S=s, A=a | Z=z_0)$ holds for all $s, a$. As a result, we can derive the identification equation for $\text{NDE}_{z_0,z_1}(R)$ as follows:
\begin{align*}
&\text{NDE}_{z_0,z_1}(R) = \sum_{s,a} \big (\mathbb{E}[R | Z=z_1, S=s, A=a]\\ - &\mathbb{E}[R | Z=z_0, S=s, A=a] \big ) P(S=s, A=a | Z=z_0),
\end{align*}
which provides a principled way to estimate $\text{NDE}_{z_0,z_1}(R)$ using standard statistical methods, hence a tool for detecting violations of dynamics fairness.
\end{proof}

\section{Algorithm for Achieving Long-Term Fairness}
\label{app2:Algorithm}
Algorithm~\ref{algo:InsightFair} presents the detailed procedure of the proposed algorithm, InsightFair, which explicitly incorporates dynamics fairness into the planning process, aiming to promote long-term fairness. Specifically, we adopt ensembles of probabilistic dynamics models to capture uncertainty in the environment, along with a planning procedure based on trajectory sampling, similar to PETS~\cite{chuaDeepReinforcementLearning2018}. However, in each epoch, the agent is required to evaluate dynamics fairness $\text{DF}$ using the proposed identification formula in Theorem~\ref{thm3:Identification} after training the augmented dynamics model $f^*_{\theta}$ on the collected data. Then it passes the value of $\text{DF}$ to the planner. 

To select the best action $a_t$ for each group in the current state $s_t$, the planner first checks whether the observed disparity $d(s_t)$ in $s_t$ is smaller than a predefined threshold $\epsilon$, which can be determined by domain experts for different applications. If $d(s_t)$ is sufficiently small and $\text{DF} = 1$, then the agent samples consistent actions for all groups. Otherwise, the agent samples group-specific action sequences for each group. Then the agent evaluates the samples using the augmented dynamics model $f^*_{\theta}$ and the estimated well-being gap $\text{TE}_{z_0,z_1}(G_{t:t+H})$ for the next $H$ steps. Instead of selecting candidate actions with the highest estimated return, to promote long-term fair well-being, the agent sorts the samples according to return regularized by the estimated well-being gap. The agent then updates the action distributions $\text{CEM}(\cdot)$ and return the first action $a_t$ in the optimal sample.

\begin{algorithm}[t] \small
    \caption{InsightFair: Algorithm for Achieving Equality in Long-Term Well-being}
    \label{algo:InsightFair}
    \begin{algorithmic}[1] 
        \Statex // Planning with Dynamics Fairness
        \Procedure{Plan}{augmented dynamics model $f^*_{\theta}$, initial state $s_t$, action distributions $\text{CEM}(\cdot)$, dynamics fairness $\text{DF}$}
            \State Compute the disparity in the current state:
            \Statex \qquad\qquad\qquad\qquad $d(s_t) = \vert s_{t,z_0} - s_{t,z_1} \vert.$
            \For{$k\in \lbrace 1,\ldots K\rbrace$} \Comment{For each iteration}
                \If{$\mathbb{I}(\text{DF})$ and $d(s_t) \leq \epsilon$}
                \State Draw $N$ samples from the same action distribution: 
                \Statex \qquad\qquad\ \ $\lbrace a_{t:t+H}^{i}\sim \text{CEM}(\cdot)\rbrace_{i=1}^{N}$
                \Else
                \State Draw $N$ samples from group-specific action
                \Statex \qquad\qquad\ \  distributions: $\lbrace a_{t:t+H}^{i}\sim \text{CEM}_{z}(\cdot)\rbrace_{i=1}^{N}$
                \Statex \qquad\qquad\ \  for each group $Z=z$.
                \EndIf
                \State Evaluate the samples using the augmented dynamics 
                \Statex \qquad\quad model $f^*_{\theta}$ and the estimated well-being gap 
                \Statex \qquad\quad $\text{TE}_{z_0,z_1}(G_{t:t+H})$.
                \State Sort the samples according to their regularized return:
                \Statex \qquad\  $\mathcal{J}(a_{t:t+H}) = \sum_{t'=t}^{t+H} \gamma^{t'-t} R_{t'} - \lambda \text{TE}_{z_0,z_1}(G_{t':t'+H}).$
                \State Update the action distributions $\text{CEM}(\cdot)$ to maximize the \Statex \qquad\quad likelihood of the top $M$ samples in $\lbrace a_{t:t+H}^{i}\rbrace_{i=1}^{N}$.
            \EndFor
            \State \textbf{return} the first action $a_t$ in the optimal sample.
        \EndProcedure
        \newline
        \Statex // Main Algorithm
        \Procedure{Learn}{initial dynamics model $f^*_{\theta_0}$, initial data set $\mathcal{D}$ collected by a random policy}
            \For{$e=1,\ldots, E$} \Comment{For each epoch}
                \State Train the augmented dynamics model $f^*_{\theta}$ on $\mathcal{D}$.
                \State Evaluate the dynamics fairness of $f^*_{\theta}$ using Theorem~\ref{thm3:Identification}.
                \State Set $\text{DF} = 0$ if the environment satisfies dynamics
                \Statex \qquad\quad fairness; otherwise, set $\text{DF} = 1$. 
                \For{$t=0,\ldots, \text{Episode Length}$} \Comment{For each episode}
                    \State Observe the current state $s_t$ from the environment.
                    \State Select actions by $a_t \leftarrow \textsc{Plan}(f^*_{\theta}, s_t, \text{CEM}(\cdot))$
                    \State Execute the action $a_t$ and observe the next state $s_{t+1}$
                    \Statex \qquad\qquad\ \ and the reward $r_t$.
                    \State Update the data set $\mathcal{D}$ with the new data point.
                \EndFor
            \EndFor
        \EndProcedure
    \end{algorithmic}
\end{algorithm}

\section{Additional Experimental Details and Results}
\label{app3：Experiments}

In this section, we provide additional details and results of the experiments presented in Section~\ref{sec4:Experiments}.

\subsection{Experimental Details of Section~\ref{subsec3.1:ExplainingInequality}}

In this experiment, to quantitatively examine the decomposition formula in Theorem~\ref{thm1:Decomposition}, we consider a simple non-linear model similar to the one studied in~\cite{mackinnonIntermediateEndpointEffect2007,zhangFairnessDecisionMakingCausal2018} but with no confounders between the treatment variable (the sensitive variable $Z$ in our case) and the outcome variable (the reward $R$ in our case). Specifically, the model with the following form:
\begin{align*}
R = \left\{\begin{array}{ll} 1, & \text{if } w_0 + w_1 z + w_2 s + w_3 a + u_{R} \geq 0\\ 0, & \text{otherwise} \end{array}\right.
\end{align*}
where $w_0, w_1, w_2, w_3$ are the model parameters and $U_{R} \sim \text{logistic}(0,1)$ is the background variable. In this case, $P(R=1 \mid Z=z, S=s, A=a)$ can be written as:
\begin{align*}
&P(R=1 \mid Z=z, S=s, A=a)\\
= &P(-w_0 > w_1 z + w_2 s + w_3 a + U_{R})\\
= &\left(1 + \exp\left(-w_0 - w_1 z - w_2 s - w_3 a\right)\right)^{-1}\\
= &L(w_0 + w_1 z + w_2 s + w_3 a),
\end{align*}
where $L(\cdot)$ is the logistic function. Assuming that the background variables $U_{S} \sim \mathcal{N}(0, \sigma_S^2)$ and $U_{A} \sim \mathcal{N}(0, \sigma_A^2)$, where $\sigma_S \ll 1,\ \sigma_A \ll 1$, we can now derive the analytical form of the natural direct and indirect effects of $Z$ on $R$ as follows:
\begin{align*}
&\text{NDE}_{z_0,z_1}(R) = \int_{-\infty}^{\infty}\int_{-\infty}^{\infty} P\left(R=1 \mid Z=z_1, S=s, A=a\right)\\
- &P\left(R=1 \mid Z=z_0, S=s, A=a\right) P(S=s,a=a \mid Z=z_0) dsda\\
= &L(w_0 + w_1) - L(w_0) + 0 (\sigma_S^2 + \sigma_A^2)
\end{align*}
\begin{align*}
&\text{NIE}_{z_1,z_0}(R) = \int_{-\infty}^{\infty}\int_{-\infty}^{\infty} P\left(R=1 \mid Z=z_1, S=s, A=a\right)\\
&\left( P(S=s,a=a \mid Z=z_0) - P(S=s,a=a \mid Z=z_1) \right) dsda\\
&= L(w_0 + w_1) - L(w_0 + w_1 + w_2 + w_3) + 0 (\sigma_S^2 + \sigma_A^2).
\end{align*}
As a result, we can compute the exact value of both $\text{NDE}_{z_0,z_1}(R)$ and $\text{NIE}_{z_1,z_0}(R)$ using these equations. To simulate the effects in different environments, we fix the model parameters and only vary $w_0$ to generate different environments. The obtained results are visualized in \figurename~\ref{fig4:Explain}. 

\subsection{Experimental Setups of Section~\ref{subsec3.2:DetectingViolation} and Section~\ref{subsec3.3:PromotingLongtermFairness}}

In these experiments, we customize the ML-fairness-gym suite~\cite{damourFairnessNotStatic2020} for group fairness problems. We conduct experiments on two environments, \emph{Allocation-v0} and \emph{Lending-v0}. Both environments are developed based on Gymnasium~\footnote{\url{https://gymnasium.farama.org/}}, which is a widely accepted python library for reinforcement learning. The studied environments use a reward vector instead of a scalar to pass group-specific rewards to the agent, with most of the other APIs remaining the same as Gymnasium. The detailed descriptions are as follows:

\emph{Allocation-v0}: In this environment, groups are characterized by different sites, each with a time-variant incident rates that determine the number of incidents (such as rat infestations) occurring at the site. The agent is tasked with allocating resources to these sites to minimize the total number of missed incidents. 
\begin{itemize}
\item[$\circ$] \textbf{Observation space}: $s_t = \left[ s_{t,z_0}, s_{t,z_1} \right]$, where $s_{t,z}$ is a scalar indicating the incident rate (mean of a Gaussian distribution) of group $z$ at time step $t$.
\item[$\circ$] \textbf{Action space}: $a_t = \left[ a_{t,z_0}, a_{t,z_1} \right]$, where $a_{t,z}$ is a two-dimensional vector indicating the allocation of resources to group $z$ at time step $t$.
\item[$\circ$] \textbf{Reward function}: For each group $z$, the reward $r_{t,z}$ is defined as the negative number of missed incidents minus the cost of allocating resources at time step $t$. Note that, the number of incidents occurred at each site is sampled from a group-specific distribution and the agent is not aware of its value. Therefore, the agent needs to take uncertainty into account when allocating resources.
\item[$\circ$] \textbf{Transition dynamics}: For each group $z$, if more than half of the incidents are solved, its incident rate will decrease by a fixed amount; otherwise, its incident rate will increase.
\end{itemize}

\emph{Lending-v0}: In this environment, each group has a dynamic credit score distribution, which determines the probability of repaying a loan. The agent is tasked with setting the loan threshold for each group to maximize the benefit of the bank. 
\begin{itemize}
\item[$\circ$] \textbf{Observation space}: $s_t = \left[ s_{t,z_0}, s_{t,z_1} \right]$, where $s_{t,z}$ is a five-dimensional vector indicating the credit score distribution (probability masses in a categorical distribution) of group $z$ at time step $t$.
\item[$\circ$] \textbf{Action space}: $a_t = \left[ a_{t,z_0}, a_{t,z_1} \right]$, where $a_{t,z}$ is a two-dimensional vector deciding the loan threshold for group $z$ at time step $t$.
\item[$\circ$] \textbf{Reward function}: For each group $z$, the reward $r_{t,z}$ is defined as the interest earned from the loans minus the cost of the loan defaults at time step $t$. Note that, in this environment, individuals with different credit scores are sampled from their group-specific distributions and this score determines whether or not they would repay the loan if granted. The agent is not aware of the true credit score of each individual and needs to take uncertainty into account when setting the loan threshold. 
\item[$\circ$] \textbf{Transition dynamics}: For each group $z$, if an individual repays the loan, a fixed amount of probability mass will be shifted from its current credit score to a higher credit score; otherwise, the probability mass will be shifted to a lower credit score.
\end{itemize}

To evaluate the effectiveness of the proposed techniques, we introduce two sets of parameters, $[\alpha_1, \alpha_2]$ and $[\beta_1, \beta_2]$, to control the relative advantage of the groups in reward assignment and state transition. Along with the initial state of each group, these parameters can be used to generate various environments in which different groups take a dominant position in turn (when the environment is unfair) or behave similarly (when the environment is fair). The parameter settings for each environment and experiment are summarized in Table~\ref{tab1:Parameters}.

\begin{table}[t]
\centering
\caption{Parameter settings for each environment and experiment.}
\label{tab1:Parameters}
\begin{tabular}{l|c|c}
\hline
\textbf{Environment} & \textbf{Allocation-v0} & \textbf{Lending-v0} \\ \hline
\multicolumn{3}{c}{Experiments in Section~\ref{subsec3.2:DetectingViolation}} \\ \hline
$s_{0,z_0}$ & 6.0 & $(0.0, 0.2, 0.3, 0.3, 0.2)$ \\
$s_{0,z_1}$ & 6.0 & $(0.0, 0.2, 0.3, 0.3, 0.2)$ \\
$\alpha$ & $[0, 1.0]$ & $[0, 0.2]$ \\
$\beta$ & $[0, 0.5]$ & $[0, 0.5]$ \\ \hline
\multicolumn{3}{c}{Experiments in Section~\ref{subsec3.3:PromotingLongtermFairness}~(unfair dynamics)} \\ \hline
$s_{0,z_0}$ & 6.0 & $(0.0, 0.2, 0.3, 0.3, 0.2)$ \\
$s_{0,z_1}$ & 6.0 & $(0.0, 0.2, 0.3, 0.3, 0.2)$ \\
$\alpha$ & 0.05 & 0.01 \\
$\beta$ & 0.05 & 0.05 \\ \hline
\multicolumn{3}{c}{Experiments in Section~\ref{subsec3.3:PromotingLongtermFairness}~(fair dynamics)} \\ \hline
$s_{0,z_0}$ & 6.2 & $(0.0, 0.2, 0.35, 0.25, 0.2)$ \\
$s_{0,z_1}$ & 6.0 & $(0.0, 0.2, 0.25, 0.35, 0.2)$ \\
$\alpha$ & 0.0 & 0.0 \\
$\beta$ & 0.0 & 0.0 \\ \hline
\end{tabular}
\end{table}

For the experiments in Section~\ref{subsec3.2:DetectingViolation}, parameters are provided in the form of intervals. We vary the parameters, $[\alpha_1, \alpha_2]$ and $[\beta_1, \beta_2]$, in these fixed intervals to generate a series of environments with different natural direct effects. For example, when $\text{NDE}_{z_0,z_1}(R)$ is being investigated, we fix $\alpha_1=\alpha_2$ and vary $\beta_1$ and $\beta_2$ in the interval $[0, 0.5]$ so that the relative advantage in state transition is zero while the relative advantage in reward assignment varies based on $\beta_2 - \beta_1$. We divide the interval $[0, 0.5]$ into $8$ equal parts so there are $8 \times 8 = 64$ environments in total for each experiment.

\subsection{Additional Experimental Results}

\onecolumn


\begin{figure}[H]
\centering
\begin{subfigure}{0.49\linewidth}
\includegraphics[width=0.99\linewidth]{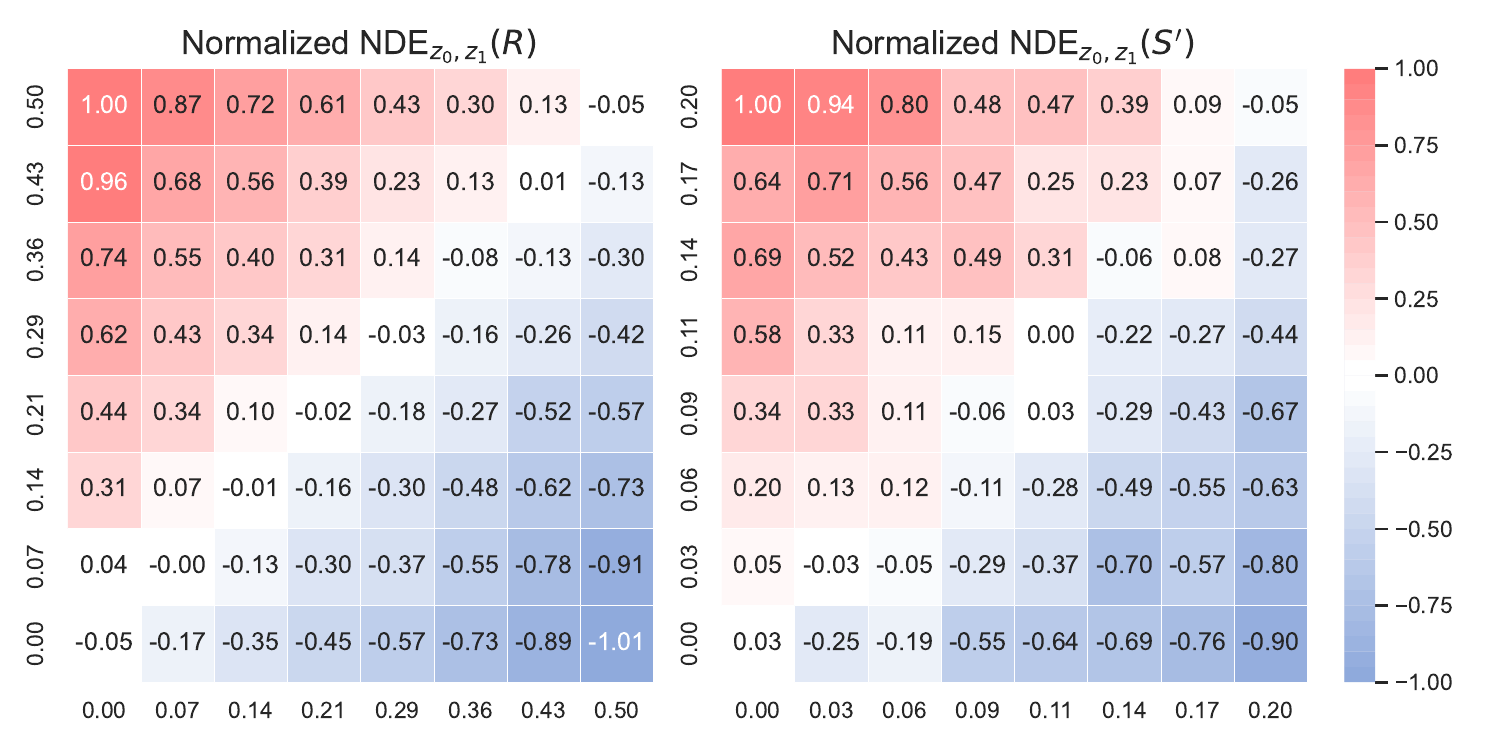} 
\end{subfigure}
\begin{subfigure}{0.49\linewidth}
\includegraphics[width=0.99\linewidth]{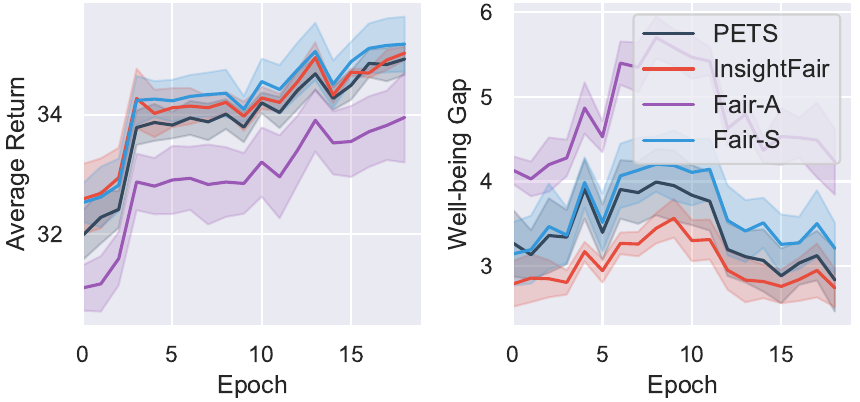}
\end{subfigure}
\vspace{-0.1in}
\caption{(Left) Results of evaluating dynamics fairness in the \emph{Lending-v0} environment. Each tile represents the estimated natural direct effects, $\text{NDE}{z_0,z_1}(R)$ or $\text{NDE}{z_0,z_1}(S')$, under specific parameter configurations indicated by the row and column. The proposed identification formula is capable of detecting the violation of dynamics fairness. (Right) Learning curves depicting the average return and the well-being gap in the \emph{Lending-v0} environment (unfair dynamics). In this environment, InsightFair achieves the smallest well-being gap with a comparable return. Fair-A performs the worst since it cannot flexibly handle the different dynamics of the two groups.}
\label{fig8:Comparison}
\end{figure}
\vspace{-0.25in}
\begin{figure}[H]
\centering
\begin{subfigure}{0.35\linewidth}
\includegraphics[width=0.99\linewidth]{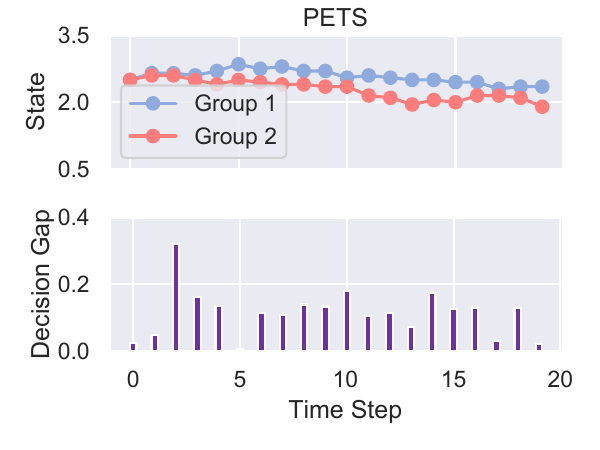} 
\end{subfigure}
\begin{subfigure}{0.35\linewidth}
\includegraphics[width=0.99\linewidth]{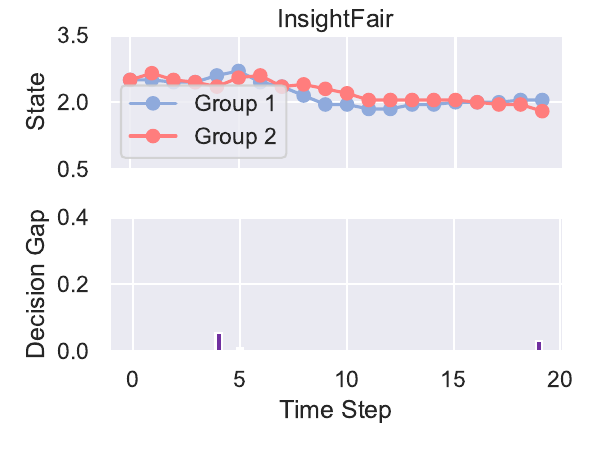}
\end{subfigure}\\
\begin{subfigure}{0.35\linewidth}
\includegraphics[width=0.99\linewidth]{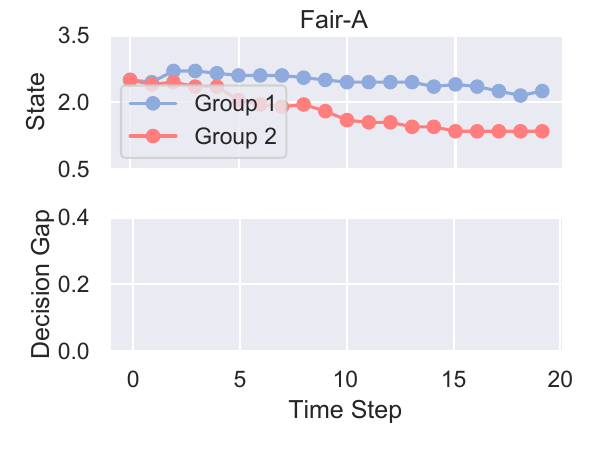} 
\end{subfigure}
\begin{subfigure}{0.35\linewidth}
\includegraphics[width=0.99\linewidth]{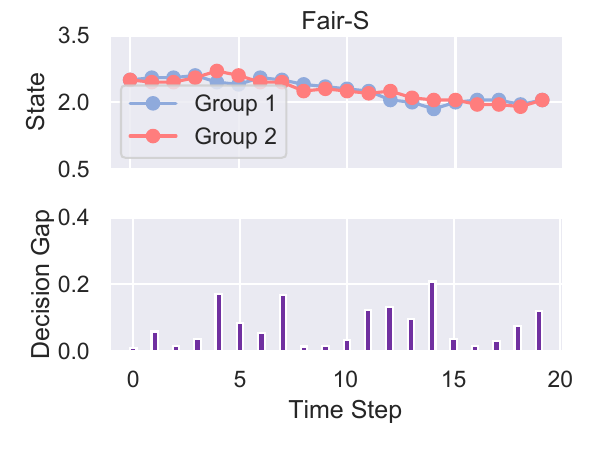}
\end{subfigure}
\vspace{-0.1in}
\caption{Changes in state and decision gaps within an episode in the \emph{Lending-v0} environment (fair dynamics). \textcolor[RGB]{255,125,125}{Red} denotes the advantaged group, which occupies a slightly better initial state as indicated in Table~\ref{tab1:Parameters}, and \textcolor[RGB]{143,170,220}{blue} denotes the disadvantaged group. InsightFair has zero decision gaps when the states (measured by the weighted sum of credit scores and its probability mass) are sufficiently close, while PETS shows large decision gaps even when the states are close. Fair-A has a zero decision gap but the state disparity increases due to the difference in initial states and the aleatoric uncertainty in the dynamics. Fair-S brings the states closer but the decision gap exists throughout the episode.}
\label{fig9:Comparison}
\end{figure}
\vspace{-0.25in}
\begin{figure}[H]
\centering
\begin{subfigure}{0.48\linewidth}
\includegraphics[width=0.99\linewidth]{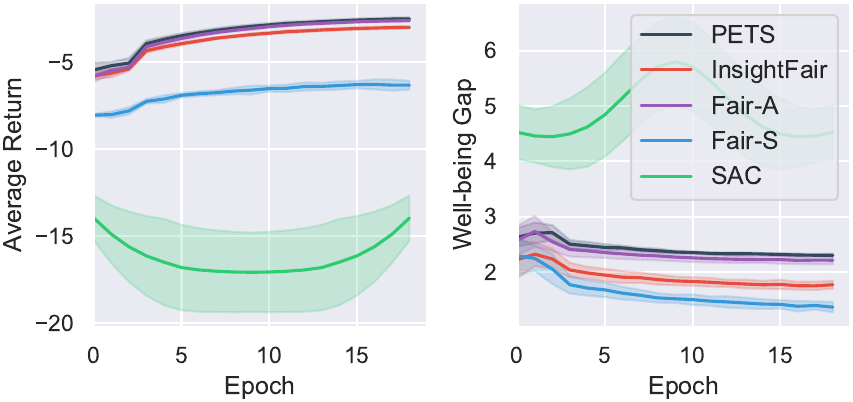} 
\end{subfigure}
\begin{subfigure}{0.48\linewidth}
\includegraphics[width=0.99\linewidth]{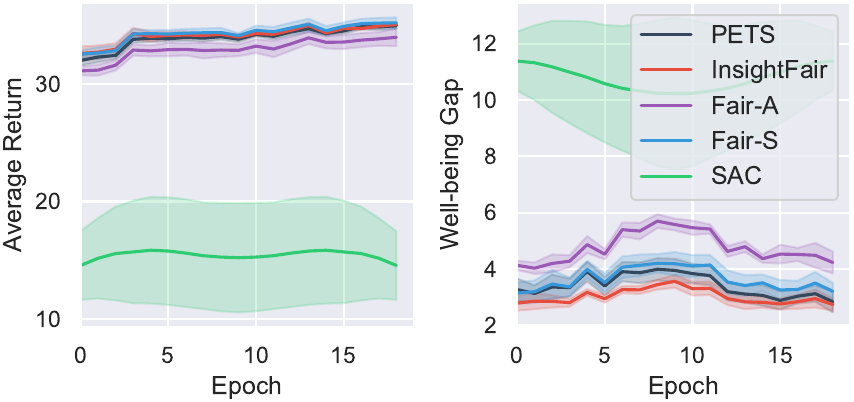}
\end{subfigure}
\vspace{-0.1in}
\caption{Additional Reuslts comparing model-based and model-free methods. Learning curves depicting the average return and the well-being gap in the \emph{Allocation-v0} (Left) and the \emph{Lending-v0} (Right) environments. The model-free method SAC consistently shows the worst performance within the training period while model-based methods benefit from the learned dynamics model and achieve better performance.}
\label{fig10:Comparison}
\end{figure}
\end{document}